\documentclass[final,3p,times]{elsarticle}

\usepackage[numbers]{natbib}
\usepackage{float}
\usepackage{amssymb}
\usepackage{amsmath,amsfonts}
\usepackage{algorithmic}
\usepackage{algorithm}
\usepackage{array}
\usepackage[caption=false,font=normalsize,labelfont=sf,textfont=sf]{subfig}
\usepackage{textcomp}
\usepackage{stfloats}
\usepackage{url}
\usepackage{verbatim}
\usepackage{multirow}
\usepackage{booktabs}
\usepackage{graphicx}
\usepackage{cite}
\usepackage{wrapfig}
\usepackage{caption}  
\usepackage{xspace}

\usepackage{amsmath}

\journal{}

\begin{document}

\begin{frontmatter}



\title{Proxy-Embedding as an Adversarial Teacher: An Embedding-Guided Bidirectional Attack for Referring Expression Segmentation Models}

\author[1]{Xingbai Chen}
\ead{chenxingbai@stu.ynu.edu.cn}

\author[2]{Tingchao Fu}
\ead{futingchao@stu.ynu.edu.cn}

\author[3]{Renyang Liu}
\ead{ryliu@nus.edu.sg}

\author[1]{Chao Yi\corref{cor1}}
\ead{yichao@ynu.edu.cn}

\author[1,2]{Wei Zhou\corref{cor1}}
\ead{zwei@ynu.edu.cn}
\cortext[cor1]{Corresponding author}


\affiliation[1]{
organization={National Pilot School of Software, Yunnan University},
    city={Kunming},
    postcode={650504}, 
    state={Yunnan},
    country={China}
}

\affiliation[2]{organization={School of Information Science and Engineering, Yunnan University},
    city={Kunming},
    postcode={650504}, 
    state={Yunnan},
    country={China}
}

\affiliation[3]{organization={Institute of Data Science, National University of Singapore},
    city={Singapore},
    postcode={117602}, 
    country={Singapore}
}

\begin{abstract}
Referring Expression Segmentation (RES) enables precise object segmentation in images based on natural language descriptions, offering high flexibility and broad applicability in real-world vision tasks. Despite its impressive performance, the robustness of RES models against adversarial examples remains largely unexplored. While prior adversarial attack methods have explored adversarial robustness on conventional segmentation models, they perform poorly when directly applied to RES models, failing to expose vulnerabilities in its multimodal structure. In practical open-world scenarios, users typically issue multiple, diverse referring expressions to interact with the same image, highlighting the need for adversarial examples that generalize across varied textual inputs. Furthermore, from the perspective of privacy protection, ensuring that RES models do not segment sensitive content without explicit authorization is a crucial aspect of enhancing the robustness and security of multimodal vision-language systems. To address these challenges, we present \textbf{PEAT}, an \textbf{Embedding-Guided Bidirectional Attack} for RES models. PEAT introduces a learnable proxy text embedding that serves as an adversarial teacher and alternates two coupled procedures during attack generation: (i) \textbf{Visual-Aligned Optimization (VAO)} that drives the image perturbation toward a target mask, and (ii) \textbf{Embedding-Guided Adversarial Optimization (EGAO)} that adversarially updates the proxy embedding to misalign the textual branch. By letting the proxy steer the visual perturbation to adapt to increasingly challenging textual inputs during optimization, PEAT produces adversarial images with strong cross-text transferability, remaining effective under unseen or semantically diverse referring expressions. Extensive experiments across multiple RES architectures and standard benchmarks show that PEAT consistently outperforms competitive baselines.
 
\end{abstract}

\begin{keyword}
Adversarial Attack, Adversarial Robustness, Referring Expression Segmentation, Model Vulnerability, Deep Learning

\end{keyword}

\end{frontmatter}



\section{Introduction}
\label{sec1}

Image segmentation is a fundamental task in computer vision, aiming to assign semantic labels to pixels and delineate object boundaries. With the development of deep learning, deep learning models~\citep{maskrcnn,fcn,deeplab} have greatly improved segmentation performance. Building on this progress, Referring Expression Segmentation (RES) models~\citep{vlt,lavt,gres,advance,3dgres,multi} enables text prompt segmentation and has shown notable success. This model combines natural language processing with visual understanding, enabling users to specify target objects in an image through natural language descriptions, thereby supporting more flexible and intelligent segmentation. Although significant progress has been made in improving segmentation accuracy and generalization in recent years, the robustness of RES models against adversarial examples~\citep{fgsm,pgd}, which are inputs perturbed with imperceptible but carefully crafted noise to mislead the model into making incorrect predictions, remains largely unexplored. While previous research \citep{segpgd,transegpgd} has extensively investigated the vulnerability of conventional segmentation models to adversarial attack, existing methods perform poorly when applied to RES models and fail to reveal the risks inherent in their unique multimodal architectures. Therefore, a comprehensive investigation into the adversarial robustness of RES models is imperative, as it enables the identification of vulnerabilities in their vision-language interaction mechanisms while also contributing to the security and reliability of multimodal systems.

\begin{figure}[!t]
\centering
\includegraphics[scale=0.4]{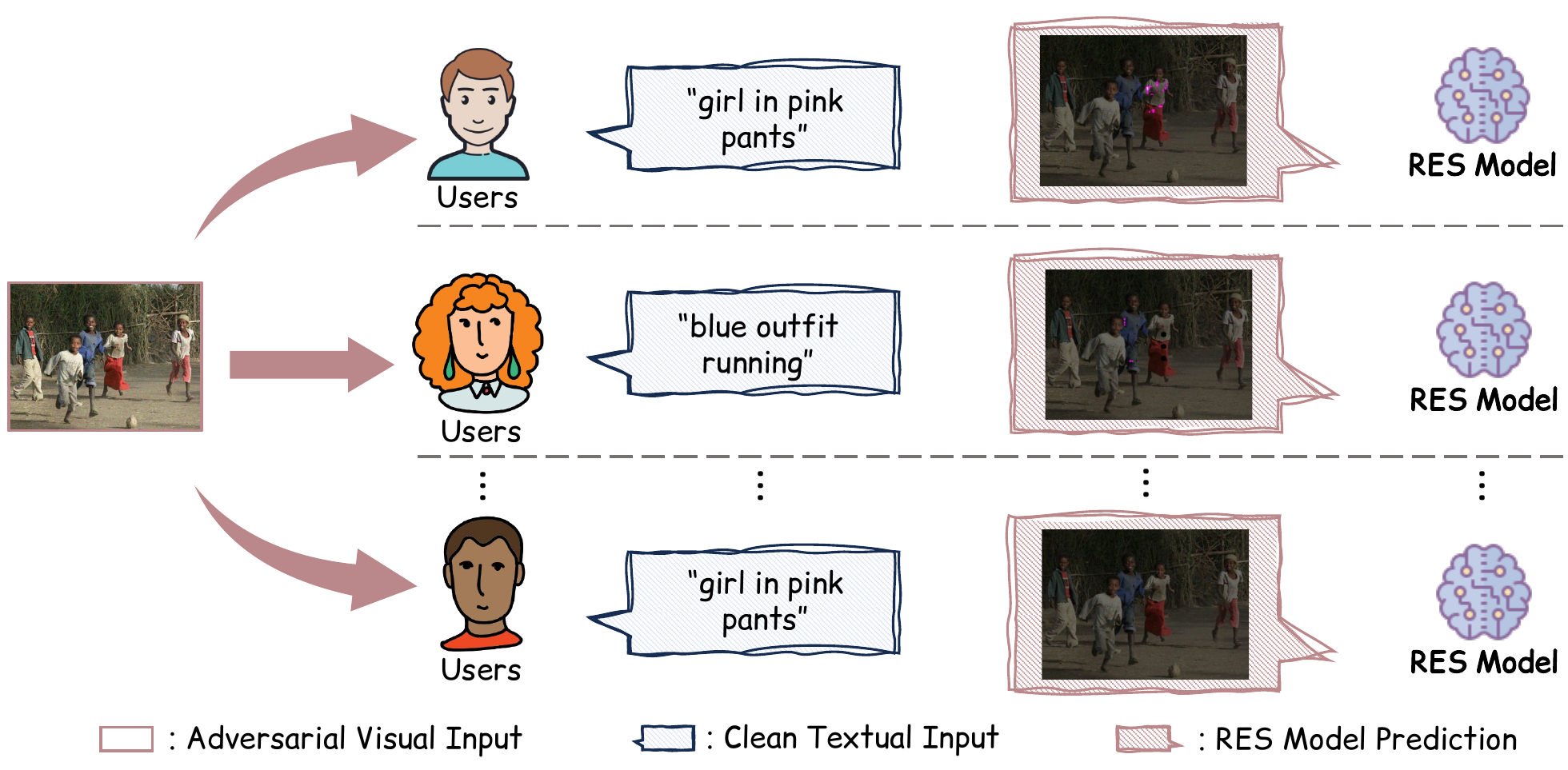}
\caption{Illustration of a cross-text transferable adversarial image generated by our method. The adversarial image is designed to mislead the RES model across a wide range of natural language descriptions provided by users. Each textual input refers to the single target object, yet the model consistently fails to generate any segmentation mask. This highlights the potential risk of adversarial examples with strong cross-text generalization in RES tasks.}
\label{fig:cross-text}
\end{figure}

Unlike conventional segmentation models~\citep{maskrcnn,unet,deeplab}, RES models inherently involve multimodal interactions between vision and language, which makes exploiting their vulnerabilities substantially more challenging. Typically, RES models take an image-text pair as input to perform target segmentation. However, it is often impossible for either attackers or defenders to anticipate how users will construct these image-text pairs in practice. This uncertainty poses substantial obstacles to the design of effective adversarial image. From the attacker's perspective, generating adversarial examples that maintain effectiveness across a broad range of textual inputs is critical for achieving consistent deception of RES models through cross-text transferability. From the defender's perspective, if adversarial images are able to consistently guide the model to produce a predefined target mask such as a full background mask across a wide range of textual inputs, thereby suppressing its segmentation capabilities. They can function as an effective defense mechanism to prevent RES models from extracting sensitive content from personal images without authorization. In summary, a desired adversarial image should maintain effectiveness under diverse and unseen textual inputs. As illustrated in Fig. \ref{fig:cross-text}, this adversarial image generated by our method exemplifies the notion of cross-text transferability in RES models. This property is particularly critical for assessing the robustness of RES models in open-world scenarios.

\begin{figure}[htbp]
\centering
\includegraphics[scale=0.7]{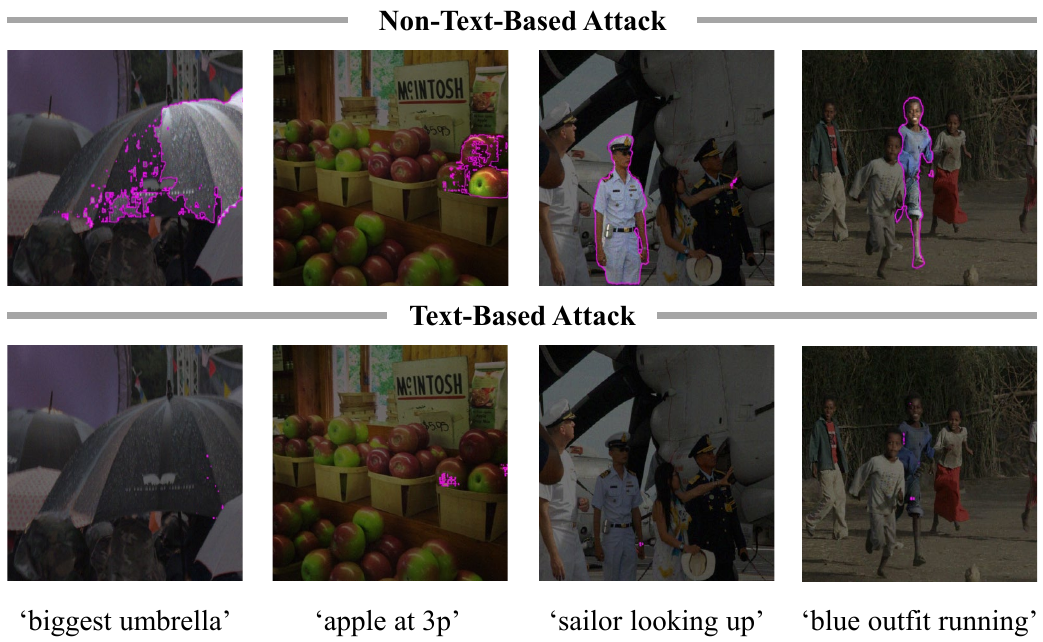}
\caption{Comparison between non-text-based and text-based adversarial attacks on RES models. The top row shows adversarial examples generated by methods that do not consider textual inputs, which fail to consistently suppress segmentation. In contrast, the middle row shows the segmentation predictions for adversarial images generated by our proposed text-based attack, which are explicitly optimized under language conditioning. The bottom row lists the corresponding textual inputs.}
\label{fig:different attack}
\end{figure}

Adversarial attacks~\citep{fgsm} are techniques that explore model robustness by misleading predictions through carefully crafted input perturbations. Although adversarial robustness has been extensively studied~\citep{segpgd,cospgd,dynamic,firstseg,attacksam,sra,psalm,transegpgd} in previous segmentation models, there are still notable limitations in the context of RES models. First, most methods~\citep{segpgd,transegpgd,psalm,cospgd,dynamic} for conventional segmentation models perturb only a single modality and do not consider the interaction with users. However, in RES tasks, both the visual modality and the textual modality jointly determine the final prediction. Representative multimodal attack methods~\citep{coattack,improvemm,sga}, such as CoAttack~\citep{coattack}, attempt to address this by perturbing both the image and the text simultaneously, generating adversarial image–text pairs. While such methods capture the multimodal nature of the task, they inherently rely on the exact paired prompts used during optimization. Consequently, this category of approaches struggles to generalize when confronted with unseen or semantically diverse referring expressions, limiting their effectiveness in open-world scenarios. Second, recent studies~\citep{sra,attacksam} targeting interactive segmentation models~\citep{sam} primarily rely on point or box prompts. As shown in Fig.~\ref{fig:different attack}, adversarial examples generated based on non-textual prompts cannot be easily transferred to RES models, and subsequent experiments have also demonstrated that they cannot generalize across different textual inputs. In summary, these approaches~\citep{attacksam,sra,segpgd,cospgd,firstseg,dynamic,transegpgd,coattack} fall short in effectively probing the adversarial robustness of RES models. Therefore, exploring the cross-text transferability of adversarial examples in RES models is of paramount importance, as it not only deepens our understanding of their multimodal robustness but also provides valuable insights for enhancing the security and reliability of such systems. that rely on natural language input.

To overcome the aforementioned limitations and systematically investigate the adversarial robustness of RES models, we propose a novel framework termed \textbf{Embedding-Guided Bidirectional Attack (PEAT)}. PEAT is designed to improve the cross-text transferability of adversarial examples by explicitly exploiting the multimodal nature of RES models. Specifically, PEAT alternates between two complementary procedures. The \textbf{Visual-Aligned Optimization (VAO)} component aligns the adversarial image with a predefined target mask by minimizing the segmentation loss, ensuring that the visual perturbation enforces a controlled prediction (e.g., a full-background mask). In parallel, the \textbf{Embedding-Guided Adversarial Optimization (EGAO)} component adversarially updates the proxy embedding to misalign the textual branch, exposing vulnerabilities in vision–language alignment and preventing overfitting to any specific textual input. By letting the proxy act as an adversarial teacher during optimization while remaining hidden at test time, PEAT generates adversarial images without the need to produce specific adversarial text, effectively extending it to a wide variety of unseen reference expressions. This bidirectional optimization strategy mirrors open-world scenarios where natural language queries are inherently unknown, thereby presenting a more realistic and potent adversarial threat. Extensive experiments across multiple RES models and benchmark datasets show that \textbf{PEAT} consistently outperforms strong baselines and retains superior effectiveness under unseen textual inputs. In brief, our contributions can be summarized as follows:
\begin{enumerate}
    \item We conduct the first comprehensive study on the adversarial robustness of Referring Expression Segmentation (RES) models, revealing their vulnerability to malicious perturbations in the multimodal setting.
    
    \item We propose a novel attack strategy named Embedding-Guided Bidirectional Attack (PEAT). This bidirectional design effectively enhances the cross-text transferability of adversarial images, enabling them to remain effective under diverse and unseen textual inputs.
    
    \item We carry out extensive experiments on multiple representative RES models and public benchmark datasets. The results demonstrate that our approach consistently maintains outstanding effectiveness and significantly outperforms existing methods.
\end{enumerate}

The remainder of this paper is organized as follows. Sec.~\ref{sec:related} reviews related work on referring expression segmentation and adversarial attacks. Sec.~\ref{sec:pre} presents the preliminary background and formulates the problem. Sec.~\ref{sec:method} details our proposed adversarial attack framework. Sec.~\ref{sec:exp} reports extensive experiments and analysis. Finally, Sec.~\ref{sec:conclu} concludes the paper and outlines future directions.

\section{Related Work}
\label{sec:related}
\subsection{Referring Expression Segmentation (RES)}
RES has emerged as a significant subtask in the field of image segmentation. As a fundamental problem in computer vision, image segmentation aims to assign semantic labels to each pixel by partitioning an image into coherent regions. The rise of deep learning has dramatically advanced this field, with seminal models such as Fully Convolutional Networks (FCN)~\citep{fcn}, Mask R-CNN~\citep{maskrcnn} and~\citep{unet,deeplab,linearformer,yang2025bi,pavani2024robust} greatly improving segmentation accuracy. More recently, foundation models have demonstrated impressive zero-shot generalization capabilities in segmentation tasks. For instance, the Segment Anything Model (SAM)~\citep{sam}, built on Vision Transformers (ViT)~\citep{vit}, supports flexible interactive segmentation through point and box prompts, without being constrained by a fixed category set.

In this context, RES has been proposed to extend conventional category-based segmentation into a language-guided setting. First introduced by~\citep{firstres}, RES allows users to specify the target object via natural language expressions, making segmentation more flexible and intuitive for users. Unlike conventional segmentation tasks that rely on predefined categories, RES enables free form object specification through human instructions. Early approaches~\citep{cnn2,cnn3} typically employed convolutional and recurrent neural networks to extract features from visual and linguistic modalities but often struggled to capture the rich cross-modality interactions. To address this limitation, later studies~\citep{graph1, graph2, graph3} incorporated graph reasoning and multimodal relation modeling, significantly enhancing the alignment between language and vision.

With the advent of large-scale vision-language pretraining~\citep{clip}, the capabilities of RES models have seen substantial improvements. Notably, \citep{evfsam} proposed EVF-SAM, which integrates a multimodal encoder \citep{beit3} into the SAM \citep{sam} architecture to support textual input segmentation, achieving state-of-the-art performance on various RES benchmarks. In parallel, DMMI~\citep{dmmi} introduces a Dual Multi-Modal Interaction framework that supports flexible expression grounding by modeling bidirectional interactions between text and vision. It effectively handles ambiguous referring expressions that may correspond to zero, one, or multiple target objects, thereby improving robustness in realistic RES scenarios. Furthermore, the rapid development of large multimodal models \citep{qwen,llava,blip,blip2,blip2} has enabled several studies \citep{sam4mllm,lisa,psalm} to leverage these pre-trained models for improved language encoding in RES, enhancing generalization in open-domain scenarios.

Despite the promising progress, the adversarial robustness of RES models remains largely unexplored. Given their increasing deployment in real-world applications such as autonomous driving \citep{drive}, medical image segmentation\citep{medical}, investigating the vulnerability of RES models under adversarial examples is not only essential for understanding their multimodal decision processes but also critical for ensuring the safety and reliability of vision-language systems.

\subsection{Adversarial Attack}

Adversarial attacks have become a central topic in the study of deep learning security. The Fast Gradient Sign Method (FGSM), introduced by~\citep{fgsm}, was the first to demonstrate that deep neural networks are highly vulnerable to small, human imperceptible perturbations intentionally crafted to cause incorrect predictions. Adversarial examples were initially introduced within the domain of image classification, where their primary objective was to deceive models into assigning incorrect labels to input samples.

Adversarial attacks are generally categorized into two types based on the attacker’s knowledge of the model: white-box and black-box attacks. In white-box attacks \citep{fgsm,pgd,liu2023aflow,cama,gradfuzz,liu2023dualflow}, the attacker has full access to the model’s architecture and parameters. In contrast, black-box attacks~\citep{bb1,bb2,bb3, MA,textguise,improving} operate under the assumption of limited or no access to the target model and primarily rely on either query-based approaches or transfer-based strategies to craft effective adversarial examples.

This techniques have also been extended to semantic segmentation tasks, where the objective shifts from misclassifying entire images to disrupting dense, pixel level predictions. One of the earliest efforts in this direction \citep{firstseg} showed that segmentation models are also vulnerable to adversarial perturbations. SegPGD~\citep{segpgd} extended the PGD~\citep{pgd} framework by distinguishing between correctly and incorrectly predicted pixels and assigning different optimization weights to these two types, improving attack efficacy. TranSegPGD~\citep{transegpgd} further advanced this idea by measuring pixel-level cross-model consistency via KL divergence~\citep{kl}, allowing the generation of adversarial examples with stronger cross-model transferability. In addition, several works~\citep{cospgd,segrobust,proximal,uncertainty,dynamic} have explored segmentation robustness from various perspectives, including multi-scale perturbations and adversarial training.

More recently, researchers have begun to investigate adversarial robustness in interactive segmentation models. Attack-SAM~\citep{attacksam} is the first attack method tailored to the Segment Anything Model (SAM)~\citep{sam}, which introduced a prompt aware loss function and generated adversarial examples guided by specified point or box prompts. However, the effectiveness of these examples is limited to the specific prompt used during optimization, and they often fail when evaluated with different prompts. To improve generalization across prompts, the SRA (Sampling-based Region Attack)~\citep{sra} approach proposed a uniform grid-based sampling strategy within the target region, enabling successful attacks regardless of the click location, thereby achieving cross-point prompt attacks Within the designated area.

Beyond point- and box-based interactive models, researchers have also explored adversarial robustness in broader multimodal vision–language models.~\citep{coattack} proposed Co-Attack, the first white-box multimodal attack that jointly perturbs both visual and textual modalities by leveraging their synergistic effects. This method highlights the vulnerability of vision–language pretraining (VLP) models when confronted with coordinated bimodal perturbations. More recently, SGA~\citep{sga} investigated black-box adversarial attacks by employing data augmentation to generate multiple groups of images and pairing them with diverse text descriptions. By comprehensively exploiting cross-modal guidance information, SGA significantly improved the transferability of adversarial examples across models in the black-box setting.

Despite these advances, existing methods exhibit limited applicability to RES models. Approaches for conventional segmentation neglect the role of textual variation, while attacks for interactive models based on point or box prompts~\citep{sam} cannot naturally extend to language-driven settings. Representative multimodal attacks such as Co-Attack~\citep{coattack} and SGA~\citep{sga} reveal vulnerabilities in vision–language models, but their effectiveness remains tied to specific image–text pairs, leading to poor generalization under unseen or semantically diverse expressions. Unlike simple point or box prompts, RES models requires resolving fine-grained cross-modal interactions between natural language and visual content. To date, no work has systematically examined the adversarial robustness of RES models under diverse textual inputs. Given their growing deployment in real-world applications, it is critical to assess these vulnerabilities and develop methods that ensure the reliability of RES systems in open-world multimodal environments.

In summary, although prior research has made notable progress in adversarial attacks on conventional and interactive segmentation models, these methods exhibit limited effectiveness when applied to RES models due to their inability to model complex language inputs. Existing segmentation attack approaches often neglect the influence of textual variation, while point- and box-based methods fail to transfer due to modality mismatches. Representative multimodal attacks such as CoAttack~\citep{coattack} and SGA~\citep{sga} further demonstrate vulnerabilities in vision–language models, but their reliance on specific paired inputs restricts their applicability to the diverse and unseen textual expressions encountered in RES models. To bridge this gap, our work introduces a novel adversarial attack framework tailored for RES models, which jointly optimizes visual and textual perturbations to systematically investigate robustness under diverse textual inputs. This represents the first comprehensive study on adversarial robustness in RES models.

\section{Preliminary}
\label{sec:pre}
\subsection{Referring Expression Segmentation Models}

RES model is a fine-grained multimodal model that aims to segment the image region corresponding to a given natural language query. Given an input image \( x_v \in \mathbb{R}^{H \times W \times 3} \) and a textual inputs \( x_t \), an RES model \( f(\cdot) \) outputs a predicted segmentation mask \( M_{pred} \in \mathbb{R}^{H \times W} \), where each value indicates the confidence that a pixel belongs to the referred object:

\begin{equation}
\mathit{M}_{pred} = f(x_v, x_t).
\end{equation}

Higher values in \( M_{pred} \) indicate a higher likelihood of inclusion in the segmented region. 

\subsection{Adversarial Attacks}

Adversarial attacks are designed to intentionally manipulate model outputs by introducing carefully crafted perturbations to the input data. In the context of target attacks, attackers aims to steer the model’s prediction toward a specific target output, rather than merely causing misclassification.

Formally, the objective of a RES model targeted attack is to find a visual modality adversarial perturbation \( \delta_v \) such that the model prediction \( f(x_v + \delta_v,x_t) \) closely approximates the predefined target mask \( {M}_{target} \), while keeping the perturbation magnitude imperceptibly small. This can be formulated as the following constrained optimization problem:

\begin{equation}
\delta_v^{*} =\underset{\delta_v}{argmin} 
\mathcal{L}_{seg}(f(x_v+\delta_v,x_t), {M}_{target}), s.t. \left\| (x_v+\delta_v) - x_v \right\|_\infty \leq \epsilon ,
\label{eqpgd}
\end{equation} 
where $\mathcal{L}_{seg}(\cdot)$ denotes a segmentation loss function, and \( \epsilon \) is the perturbation budget that imposes an \( \ell_\infty \)-norm constraint on the perturbation, limiting the maximum allowable change to any individual pixel. The optimized perturbation \( \delta_v^{*} \) represents the adversarial noise added to the visual input that most effectively drives the model output toward the target mask \( M_{target} \) under the given constraint.





\subsection{Problem Formulation}

In this work, we aim to craft a single visual perturbation $\delta_v$ that can universally mislead the RES model across a set of diverse textual inputs. Let $\mathcal{X}_t = \{x_t^1, x_t^2, \dots, x_t^k\}$ denote a textual inputs set for evaluating cross-text transferability and attack effectiveness. For a clean image $x_v$, our objective is to generate an adversarial image $x_v + \delta_v$ such that, when paired with any textual input $x_t^k \in \mathcal{X}_t$, the RES model consistently outputs a predefined target mask $M_{target}$ (e.g., an all-background mask):

\begin{equation}
M'_{pred} = f(x_v + \delta_v, x_t^1) = f(x_v + \delta_v, x_t^2) = \cdots = f(x_v + \delta_v, x_t^k) \approx M_{target},
\end{equation}
where, $M'_{pred}$ denotes the predicted segmentation mask generated from the adversarial image, and $\delta_v$ is the visual perturbation optimized to deceive the model under all provided text inputs, as detailed in Eq.~\ref{eqpgd}.

\section{Methodology}
\label{sec:method}
In this section, we introduce the threat model and our detailed attack method. We start with a basic Single-Text attack setting, then extend it to a Multi-Text attack. Finally, we leverage \textbf{Embedding-Guided Bidirectional Attack (PEAT)} to further enhance cross-text transferability.

\subsection{Basic Attack Strategies: Single and Multi-Text Attacks}
A simple strategy for generating adversarial examples against RES models involves optimizing image perturbations under the guidance of a single textual input; this approach is referred to as Single-Text. Building upon this, to further enhance the cross-text transferability of perturbations, a direct approach is to utilize multiple textual inputs when updating image perturbations, which is referred to as Multi-Text.

The objective of the attack is to generate a visual perturbation \( \delta_v \) based on these textual inputs to ensure that, for single or multiple data pairs from \(\mathcal{X}_t\), the model produces the predefined target mask \( M_{\text{target}} \). This optimization can be mathematically represented as:
\begin{equation}
\small
        \underset{\delta_v}{min} \sum_{k=1}^{K} \mathcal{L}_{seg}(f(x_v + \delta_v, x_t^k), M_{target}).
\end{equation}

The distinction between Single-Text and Multi-Text lies in whether \( K \) equals 1. Specifically, in the Single-Text setting, the perturbation is optimized using a single textual input (\( K = 1 \)), whereas in the Multi-Text setting, the optimization is performed across multiple textual inputs (\( K > 1 \)), with the aim of enhancing the cross-text transferability of the adversarial perturbation.
\begin{figure}[t]
    \centering
    \includegraphics[width=\linewidth]{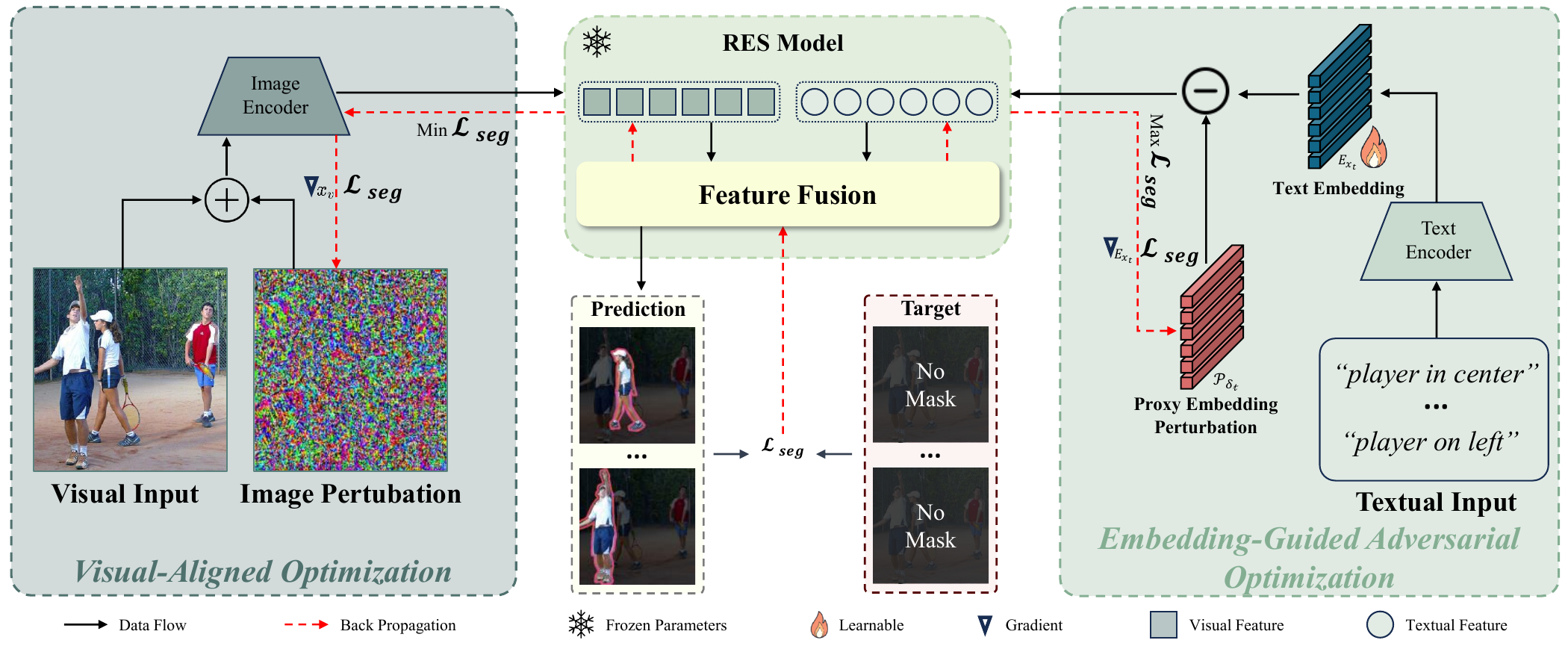}
    \caption[Overview of PEAT framework]
    {\footnotesize{
    Overview of \textbf{PEAT} framework. Both the image perturbation $\delta_v$ and the proxy embedding perturbation $\mathcal{P}_{x_t}$ are learnable, but the proxy perturbation does not collaborate with $\delta_v$ to deceive the model. They are optimize with opposite goals: $\delta_v$ aims to minimize the segmentation loss, while $\mathcal{P}_{x_t}$ aims to maximize the mask loss. The update frequency of the image perturbation and proxy perturbation can be different.
    }}
    \label{fig:framework}
\end{figure}

However, we find that adversarial images generated by this basic attack method are still likely to be correctly segmented by the model when encountering unseen textual inputs. Essentially, these basic methods tend to overfit the textual inputs involved in the adversarial images generation process.

\subsection{Embedding-Guided Bidirectional Attack (PEAT)}
To alleviate overfitting to specific texts during adversarial image generation and to improve the transferability of adversarial images across diverse textual inputs, we propose a novel attack framework called \textbf{Embedding-Guided Bidirectional Attack (PEAT)}. Built on the multimodal nature of RES models, PEAT adopts a cooperative optimization strategy that alternates between two complementary branches: \textbf{Visual-Aligned Optimization} (VAO) for the visual modality and \textbf{Embedding-Guided Adversarial Optimization} (EGAO) for the textual modality. Concretely, the VAO branch minimizes the segmentation loss between the model's predicted mask and a predefined target mask, thereby guiding the image perturbation toward a desired prediction (e.g., a full-background mask). In the EGAO branch, we introduce a learnable proxy embedding perturbation as a continuous surrogate for discrete textual inputs; EGAO adversarially updates the proxy embedding to misalign the textual branch, simulating challenging and diverse language cues. Crucially, the proxy embedding does more than merely stress the language branch: by producing adversarial directions in embedding space that are semantically meaningful (rather than token-specific), it provides a cross-modal guidance signal that steers the visual perturbation toward failure modes that are invariant to paraphrase and other linguistic variations. In this way, the proxy acts as an adversarial teacher that actively directs the image perturbation to develop more generalizable (cross-text) attack patterns, reducing text-specific overfitting. By alternately optimizing the visual and embedding-guided textual branches, PEAT refines image perturbations collaboratively in the joint vision–language space, producing adversarial images that generalize well and maintain strong attack effectiveness under unseen or semantically varied referring expressions. Implementation details for each component are provided in the following sections.

\subsubsection{Visual-Aligned Optimization}

In the Visual-Aligned Optimization phase, our primary objective is to iteratively update the adversarial image so that the output mask produced by the RES model, given a set of textual inputs, closely aligns with a predefined target mask such as a full background. This phase is conceptually consistent with adversarial optimization in Multi-Text settings.

Specifically, while keeping the textual inputs fixed, we perform gradient descent optimization on the image to minimize the segmentation loss between the predicted mask and the target mask $M_{target}$. The update rule for the adversarial image ${x'}_v$ at iteration $i+1$ is defined as:

\begin{equation}
{x'}_v^{(i+1)} = \Pi_{\epsilon_{v},x_v}\left( {x'}_v^{(i)} - \alpha_v \cdot sign \left( \nabla_{x_v}\sum_{k=1}^{K} \mathcal{L}_{seg}\left(f\left({x'}_v^{(i)} , \mathbf{E}^{(k)}_{x_t}\right), M_{target}\right) \right) \right),
\end{equation}
where $\Pi_{\epsilon_{v},x_v}(\cdot)$ denotes a projection operation that ensures the updated image remains within an $\epsilon_v$-bounded neighborhood of the original clean image $x_v$; $\alpha_v$ is the visual perturbation step size; the specific loss function $\mathcal{L}_{seg}(\cdot)$ will be introduced in Sec.~\ref{loss_sec}, and $\{\mathbf{E}^{(k)}_{x_t}\}_{k=1}^{K}$ denotes the $k_{th}$ text embedding of textual inputs. The gradient is aggregated over all texts to encourage visual perturbations that are semantically aligned across diverse textual inputs.

The resulting visual perturbation $\delta_v$ at iteration $i+1$ can be expressed as:

\begin{equation}
\delta_v^{(i)} = \alpha_v \cdot \text{sign} \left( \nabla_{x_v}\sum_{k=1}^{K} \mathcal{L}_{seg}\left(f({x'}_v^{(i)} , \mathbf{E}^{(k)}_{x_t}), M_{\text{target}}\right) \right).
\end{equation}

This process can be viewed as a minimization objective over the visual modality:

\begin{equation}
\min_{\delta_v} \sum_{k=1}^{K}\mathcal{L}_{seg}\left(f\left(x_v + \delta_v, \mathbf{E}^{(k)}_{x_t}\right), M_{target}\right).
\end{equation}

This optimization strategy progressively aligns the visual perturbation $\delta_v$ with the semantic structure of the target mask, thereby enabling the adversarial image to achieve strong attack effectiveness under fixed textual inputs. This alignment in the visual modality serves as a solid foundation for the subsequent adversarial optimization within the language modality.

\subsubsection{Embedding-Guided Adversarial Optimization (EGAO)}

To further enhance the cross-text transferability of adversarial images, we introduce \textbf{Embedding-Guided Adversarial Optimization} (EGAO). Unlike the visual modality, which naturally lies in a continuous space, textual inputs are inherently discrete sequences, making it impossible to directly backpropagate gradients through the original tokenized input. To overcome this limitation, we introduce a proxy embedding perturbation $\mathcal{P}_{\delta_t}$ applied in the embedding space. This proxy embedding serves as a continuous surrogate for discrete textual inputs, enabling gradient-based adversarial updates. For brevity, we refer to it as proxy embedding perturbation in the remainder of this paper. Importantly, the proxy embedding perturbation is used only during the optimization stage and is never part of the final adversarial input; instead, its role is to act as an adversarial teacher that guides the generation of visual perturbations.

The objective of this branch is to maximize the segmentation loss between the model’s prediction and a predefined target mask (e.g., a full background) by adversarially updating the proxy embedding to misalign the textual branch. By periodically injecting these adversarial updates, the proxy embedding exposes vulnerabilities in the vision–language alignment and simulates more diverse user expressions. Crucially, this guided optimization process drives the visual perturbation to induce text-invariant segmentation breakdowns that remain effective even when evaluated against unseen or semantically varied textual inputs.

The update rule for the adversarial proxy embedding $\widetilde{\mathbf{E}}^{(k)(i+1)}_{x_t}$ of the $k^{th}$ textual input at iteration $i+1$ is defined as:
\begin{equation}
\widetilde{\mathbf{E}}^{(k)(i+1)}_{x_t} = \Pi_{\epsilon_{t},\mathbf{E}_{x_t}} \left( \widetilde{\mathbf{E}}^{(k)(i)}_{x_t} + \alpha_t \cdot \text{sign} \left( \nabla_{\mathbf{E}^{(k)}_{x_t}} \mathcal{L}_{seg}\left(f\left(x_v, \widetilde{\mathbf{E}}^{(k)(i)}_{x_t}\right), M_{target}\right) \right) \right),
\end{equation}
where $\alpha_t$ is the step size and $\Pi_{\epsilon_t, \mathbf{E}_{x_t}}(\cdot)$ denotes the projection operator that constrains the adversarial embedding within an $\epsilon_t$-bounded $\ell_\infty$ ball around the clean embedding.

The resulting proxy embedding perturbation $\mathcal{P}_{\delta_t}^{(k)(i)}$ for the $k^{th}$ textual input at iteration $i+1$ can be expressed as:
\begin{equation}
\mathcal{P}_{\delta_t}^{(k)(i)} = \alpha_t \cdot \text{sign} \left( \nabla_{\mathbf{E}^{(k)}_{x_t}} \mathcal{L}_{seg}(f(x_v, \mathbf{E}^{(k)(i)}_{x_t}), M_{target}) \right).
\end{equation}

This maximization step explicitly pushes the proxy embedding in the opposite direction of Visual-Aligned Optimization, thereby providing adversarial guidance that steers the visual perturbation toward more generalizable and cross-text robust solutions. Overall, the process can be summarized as solving the following maximization objective:
\begin{equation}
\max_{\mathcal{P}_{\delta_t}} \sum_{k=1}^{K}\mathcal{L}_{seg}\left(f\left(x_v, \mathbf{E}^{(k)}_{x_t} + \mathcal{P}^{(k)}_{\delta_t}\right), M_{target}\right).
\end{equation}

Through the EGAO branch, we implement an embedding-guided adversarial intervention that not only prevents overfitting to specific textual inputs but also actively directs visual perturbations toward semantically invariant vulnerabilities. When integrated with the VAO branch, this strategy forms a cohesive framework (PEAT) capable of generating adversarial images that are both highly transferable and semantically robust across diverse textual inputs.

\subsubsection{Joint Bidirectional Optimization}

To jointly optimize perturbations in both visual and textual modalities, we propose a unified bidirectional optimization strategy under the PEAT framework. PEAT integrates the Visual-Aligned Optimization (VAO) and Embedding-Guided Adversarial Optimization (EGAO) branches into a single iterative min–max loop, aiming to produce adversarial images that both conform to a predefined target mask and generalize across diverse, unseen textual inputs.

The overall pipeline is illustrated in Fig.~\ref{fig:framework}. At each iteration, PEAT first holds the textual input fixed and performs gradient-descent updates on the image perturbation $\delta_v$ under the VAO objective, minimizing the segmentation loss between the model prediction and the target mask (e.g., a full background mask). This alignment-oriented optimization drives the visual perturbation toward the intended attack goal. Periodically, and according to a predetermined schedule, PEAT activates the EGAO branch: it adversarially updates the proxy embedding perturbation to misalign the textual branch, thereby exposing semantic failure modes in the vision–language alignment and simulating a broader distribution of user expressions. (We use the term proxy embedding perturbation for this construct throughout the paper; it is employed only during optimization and is not included in final test-time inputs.)

We cast the combined procedure as a min–max optimization:
\begin{equation}
\min_{\delta_v}\; \max_{\mathcal{P}_{\delta_t}} \; 
\mathcal{L}_{seg}\Big( f\big(x_v + \delta_v,\; \mathbf{E}_{x_t} + \mathcal{P}_{\delta_t}\cdot \mathbb{I}_{[i \bmod N = 0]}\big),\; M_{target} \Big),
\end{equation}
where $\mathbb{I}_{[i \bmod N = 0]}$ is an indicator controlling the periodic activation of EGAO (every $N$ visual steps). The projection operators and norm budgets for $\delta_v$ and $\mathcal{P}_{\delta_t}$ are applied as in the single-branch descriptions to ensure perturbations remain within prescribed bounds.

A key design choice is the update frequency $N$ of the proxy embedding perturbation. Updating too frequently can destabilize the VAO alignment and harm immediate attack effectiveness; updating too rarely reduces the simulated linguistic diversity and weakens cross-text transferability. We therefore perform one adversarial proxy update every $N$ visual steps and empirically study the trade-off and sensitivity of $N$ in Sec.~\ref{interval}.

By alternately optimizing both modalities in this min–max fashion, PEAT effectively searches for robust adversarial directions in the joint vision–language space. The resulting adversarial images consistently mislead RES models across a wide range of textual expressions, thereby exposing critical vulnerabilities in multimodal reasoning. The full optimization routine is summarized in Algorithm~\ref{algo:algorithm}.

\begin{algorithm}[t]
\footnotesize
\caption{PEAT: Embedding-Guided Bidirectional Attack}
\label{algo:algorithm}
\begin{algorithmic}[1]
\REQUIRE RES Model \(f\), target Mask $M_{target}$, visual input \(x_v\), textual input \(x_t^k\) \(\in\) text set \(\mathcal{X}_t = \{x_t^1, x_t^2, \dots, x_t^k\}\), image perturbation budget \(\epsilon_v\), proxy embedding perturbation budget \(\epsilon_t\), step size of images perturbation updating \(\alpha_v\),  step size of proxy embedding perturbation updating \(\alpha_t\), number of iteration steps \(I\), proxy embedding perturbation update interval \(N\)
\ENSURE Adversarial image \(x_v'\)
\STATE Initialize adversarial image: \(x_v' = x_v\)
\STATE Initialize adversarial text embedding:
\FOR{\(k\) = 1 to \(K\)}
\STATE \(\widetilde{\mathbf{E}}^{k}_{x_t} = \text{Text Encoder}(x^{k}_t)\)
\ENDFOR
\FOR{step = 1 to \(I\)}
    \STATE Compute gradient for adversarial image: \(g_v =   \nabla_{x_v}\sum_{k=1}^{K} \mathcal{L}_{seg}(f(x_v' , \widetilde{\mathbf{E}}^{k}_{x_t}), M_{target})\)
    \STATE Update with gradient descent: \(x_v' = x_v' - \alpha_v \cdot sign(g_v)\)
    \IF{mod(step, N) == 0}
        \FOR{\(k\) = 1 to \(K\)}
            \STATE Compute gradient for adversarial text embedding: \(g_{{\mathbf{E}}^{k}_{x_t}} =  \nabla_{{\mathbf{E}}^{k}_{x_t}} \mathcal{L}_{seg}(f(x_v' , \widetilde{\mathbf{E}}^{k}_{x_t}), M_{target})\)
            \STATE Update with gradient ascent: \(\widetilde{\mathbf{E}}^k_{x_t} =\widetilde{\mathbf{E}}^k_{x_t} + \alpha_t \cdot \text{sign}(g_{{\mathbf{E}}^{k}_{x_t}})\)
            \STATE Project \(\widetilde{\mathbf{E}}^k_{x_t}\) to be within the \(\epsilon\)-ball of \({\mathbf{E}}^k_{x_t}\): \(\widetilde{\mathbf{E}}^k_{x_t} = \Pi_{\epsilon_{t},\mathbf{E}_{x_t}}(\widetilde{\mathbf{E}}^k_{x_t})\)
        \ENDFOR
    \ENDIF
    \STATE Project \(x_v'\) to be within the \(\epsilon\)-ball of \(x_v\): \(x_v' = \Pi_{\epsilon_{v},x_v}(x_v')\)
\ENDFOR
\RETURN \(x_v'\)
\end{algorithmic}
\end{algorithm}

\subsection{Loss Function}
\label{loss_sec}
In SAM-based RES models~\citep{evfsam}, the SAM~\citep{sam} outputs positive values for foreground pixels and negative values for background pixels. Following the loss function formulation in Attack-SAM~\citep{attacksam}, we omit gradient computation for pixels that are already predicted as negative. Instead, we focus on those pixels predicted with positive values. The loss function for SAM-based RES models is defined as follows: 
\begin{equation}
\mathcal{L}_{seg}(f(x_v, x_t), M_{target}) = \left\|Clip(f(x_v, x_t),min = Neg_{th}) - Neg_{th}\right\|^2,
\end{equation}
where \(\mathcal{L}_{seg} \) represents the Mean Squared Error (MSE) loss function, and \( Clip \) ensures that pixels with values lower than the threshold \( Neg_{th} \) have zero gradients. This is because computing gradients for pixels that are already not segmented is meaningless. Following Attack-SAM, we set \( Neg_{th} \) to -10.

In non-SAM-based RES models~\citep{dmmi, lavt}, we adopt the cross-entropy loss function used during model~\citep{dmmi} training as the attack loss function. The weight of the cross-entropy loss remains the same as that used in the model~\citep{dmmi} training process, specifically set to (0.9, 1.1). The loss function for non-SAM-based RES models is defined as:

\begin{equation}
\mathcal{L}_{seg}(f(x_v, x_t), M_{target}) = w_0 \cdot \mathcal{L}_{CE}(f_0(x_v, x_t), M_{target} ) + w_1 \cdot \mathcal{L}_{CE}(f_1(x_v, x_t), M_{target}),
\end{equation}
where \( f_i(x_v, x_t) \) denotes the model prediction logits for class \( i \in \{0,1\} \), and \( M_{\text{target}} = i \) extracts the spatial positions where the ground-truth label equals \( i \). \( w_0 = 0.9 \) and \( w_1 = 1.1 \) are the class weights for background and foreground, respectively.

\section{Experiment}
\label{sec:exp}
\subsection{Experimental Setup}


\textbf{Datasets}. We evaluate our method on three benchmark datasets: RefCOCO~\citep{refcoco}, RefCOCO+~\citep{refcoco+}, and RefCOCOg~\citep{refcocog}, which are widely used in referring expression comprehension and provide rich image–text pairs. RefCOCO is derived from MSCOCO~\citep{mscoco} and contains 142,209 expressions across 50,000 images, collected in an interactive setting with relatively short descriptions. RefCOCO+ is constructed in a similar manner but excludes absolute positional words (e.g., ``on the right''), while RefCOCOg features longer and more descriptive expressions, making it suitable for testing complex vision–language reasoning. Following the dataset setting in~\citep{evfsam}, we select the first 750 images from each dataset for evaluation. To assess cross-text transferability, we divide the referring expressions into an optimization phase (train set) and an evaluation phase (test set). In the single-text setting, one referring expression per target object is used for optimization, whereas in the multi-text setting, multiple expressions per object are included. The remaining expressions are reserved for evaluation, ensuring no overlap between optimization and evaluation inputs. This setup provides a reliable protocol for measuring attack transferability, with the ratio of optimization to evaluation expressions kept at approximately 1:4.

\textbf{Metrics}. We use three complementary metrics, among which ASR and mIoU-UTM are proposed in this work to specifically evaluate the suppression of segmentation capability.

\emph{Attack Success Rate (ASR, \%)}. For sample $i$, let $S_i$ be the predicted foreground mask under the adversarial image and $G_i$ the ground-truth foreground mask. We denote $|\cdot|$ as the number of pixels in a mask. The mask ratio is defined as
\begin{equation}
r_i \;=\; \frac{|S_i|}{|G_i|}.
\end{equation}
An attack is considered successful if $r_i \le \tau$, where $\tau$ is a predefined threshold. The overall rate is
\begin{equation}
\text{ASR@}\tau \;=\; \frac{1}{N}\sum_{i=1}^{N} \mathbb{I}[\,r_i \le \tau\,],
\end{equation}
where $\mathbb{I}[\cdot]$ is the indicator function. As a metric introduced in this paper, ASR directly measures how effectively the adversarial perturbation reduces the segmented area relative to the ground truth, thereby reflecting suppression strength.

\emph{Mean IoU with Ground Truth (mIoU-GT, \%)}. This standard metric measures the overlap between adversarial prediction $S_i$ and ground truth $G_i$:
\begin{equation}
\text{mIoU-GT} \;=\; \frac{1}{N}\sum_{i=1}^{N} \frac{|S_i \cap G_i|}{|S_i \cup G_i|}.
\end{equation}
Lower values indicate stronger deviation from ground truth, i.e., a more effective attack.

\emph{Mean IoU with Unsegmented Target Mask (mIoU-UTM, \%)}. Let $B_i$ denote the background (unsegmented pixels) of $S_i$, and $M_{\text{target}}$ the predefined target mask (set as full background). We compute
\begin{equation}
\text{mIoU-UTM} \;=\; \frac{1}{N}\sum_{i=1}^{N} \frac{|B_i \cap M_{\text{target}}|}{|B_i \cup M_{\text{target}}|}.
\end{equation}
Proposed in this work, mIoU-UTM quantifies how well the unsegmented regions align with the target background. A higher value indicates that the adversarial perturbation more effectively suppresses the model’s segmentation ability.

Together, these three metrics comprehensively evaluate both conventional segmentation accuracy degradation (via mIoU-GT) and the proposed suppression oriented objectives (via ASR and mIoU-UTM), enabling a more thorough analysis of attack effectiveness and cross-text transferability.

\textbf{Models}. We conduct experiments on five representative RES models: EVF-SAM~\citep{evfsam}, EVF-SAM2~\citep{evfsam}, DMMI~\citep{dmmi} with both ResNet~\citep{resnet} and Swin Transformer~\citep{swin} backbones, and LAVT~\citep{lavt} based on a Swin Transformer~\citep{swin} backbone. These models collectively reflect state-of-the-art architectures for RES, incorporating strong multimodal encoders and text-guided mask decoders. Our selection covers a broad spectrum of design paradigms: large vision foundation models~\citep{sam,sam2} (SAM family), convolutional architectures (ResNet backbone), and transformer-based architectures (SwinT backbone). To comprehensively evaluate adversarial robustness, we adopt a white-box attack setting across all models, assuming full access to model architectures and parameters. This diverse set of models has also been widely adopted as benchmarks for assessing RES models performance and robustness in prior work, ensuring the generality of our evaluation.

\textbf{Baselines}. Since there is currently no dedicated adversarial attack research against RES models, we adapt existing methods for segmentation models to fit RES models. For SAM-based RES models (EVF-SAM, EVF-SAM2), we select Attack-SAM~\citep{attacksam}, SRA~\citep{sra}, SegPGD-Single, SegPGD-Multi, Single-Text, and Multi-Text as baseline methods. Notably, since Attack-SAM and SRA generate adversarial samples based on SAM’s point prompt mechanism, they are not applicable to non-SAM RES models such as DMMI~\citep{dmmi,lavt} and are thus excluded from comparisons on DMMI and LAVT. Specifically, SegPGD-Single employs a single textual input to guide the SegPGD~\citep{segpgd} algorithm in optimizing adversarial images, while SegPGD-Multi leverages multiple textual inputs for optimization. Single-Text and SegPGD-Single share the same textual input with identical numbers. Similarly, Multi-Text, SegPGD-Multi, and our proposed method all utilize multiple textual inputs during optimization, ensuring that they use the same textual inputs and maintain consistency in the number of texts for a fair comparison. 

In addition, given the multimodal interaction nature of RES models, we also compare with CoAttack~\citep{coattack}, a representative multimodal attack method. CoAttack generates adversarial image–text pairs by jointly perturbing both modalities, which differs from our RES setting that emphasizes cross-text transferability under unseen expressions. To adapt CoAttack for our evaluation, we only use the adversarial images it produces as test inputs, discarding the paired adversarial text. Although the application scenarios are not fully aligned, this comparison provides valuable insight into how a strong multimodal baseline performs when applied to RES models, thereby highlighting the distinctiveness and advantages of our proposed PEAT framework.

\textbf{Implementation}. All experiments are performed on an NVIDIA A100 GPU to ensure computational efficiency and scalability. Adversarial images are generated by applying perturbations to both the visual input and text embedding. For the visual modality, the perturbation step size is set to $2/255$, with a maximum $\ell_\infty$ constraint of $16/255$. The attack is optimized for 100 iterations. Notably, the textual perturbation is introduced only during the optimization phase; during testing, only the adversarial image is used as input to evaluate the attack effectiveness under diverse textual inputs. For the proxy embedding perturbation, the perturbation is applied directly to the embedding with a step size of $0.01$ and a maximum budget of $0.2$. The choice of proxy embedding perturbation step size is empirically validated in the ablation studies presented in Sec.~\ref{step}.

\subsection{Overall Performance}

\begin{table*}[!t]
\centering
\resizebox{\textwidth}{!}{
\begin{tabular}{l l ccc ccc ccc}
\toprule
\textbf{Model} & \textbf{Method} & \multicolumn{3}{c}{\textbf{RefCOCO}} & \multicolumn{3}{c}{\textbf{RefCOCOg}} & \multicolumn{3}{c}{\textbf{RefCOCO+}} \\
\cmidrule(lr){3-5} \cmidrule(lr){6-8} \cmidrule(lr){9-11}
& & ASR@30$\uparrow$ & mIoU-GT$\downarrow$ & mIoU-UTM$\uparrow$
  & ASR@30$\uparrow$ & mIoU-GT$\downarrow$ & mIoU-UTM$\uparrow$
  & ASR@30$\uparrow$ & mIoU-GT$\downarrow$ & mIoU-UTM$\uparrow$ \\
\midrule

\multirow{8}{*}{EVF-SAM}
& Attack-SAM      & 5.43  & 76.99 & 88.95 & 1.65  & 68.85 & 91.51 & 1.11  & 72.89 & 92.03 \\
& SRA             & 5.98  & 74.96 & 89.25 & 1.57  & 67.57 & 91.79 & 1.19  & 71.23 & 92.29 \\
& Single-Text     & 36.61 & 40.17 & 93.79 & 19.07 & 48.65 & 93.04 & 18.43 & 50.22 & 93.75 \\
& Multi-Text      & 81.72 & 10.79 & 98.49 & 77.00 & 12.47 & 98.25 & 75.26 & 14.61 & 98.41 \\
& SegPGD-Single   & 34.76 & 41.57 & 93.68 & 18.66 & 48.46 & 93.23 & 17.42 & 50.84 & 93.76 \\
& SegPGD-Multi    & 45.14 & 28.44 & \textbf{98.70} & 44.42 & 26.10 & 96.29 & 40.07 & 28.94 & 96.45 \\
& CoAttack        & 2.56    & 50.10    & 83.46    & 3.84    & 42.17    & 84.66    & 2.94    & 48.98    & 85.32    \\
& \textbf{Ours}   & \textbf{84.06} & \textbf{9.85} & 98.65 & \textbf{79.12} & \textbf{11.27} & \textbf{98.41} & \textbf{77.12} & \textbf{13.43} & \textbf{98.52} \\
\midrule

\multirow{8}{*}{EVF-SAM2}
& Attack-SAM      & 0.79  & 77.13 & 88.72 & 2.13  & 71.19 & 91.23 & 1.60  & 73.82 & 91.80 \\
& SRA             & 0.68  & 77.72 & 88.78 & 1.46  & 72.07 & 91.17 & 1.01  & 74.45 & 91.81 \\
& Single-Text     & 15.97 & 60.45 & 91.44 & 12.00 & 61.25 & 92.48 & 10.25 & 64.23 & 92.86 \\
& Multi-Text      & 26.39 & 47.29 & 94.01 & 23.97 & 45.62 & 94.87 & 25.01 & 47.05 & 95.22 \\
& SegPGD-Single   & 16.89 & 60.19 & 91.61 & 12.04 & 61.37 & 92.40 & 10.49 & 64.65 & 92.89 \\
& SegPGD-Multi    & 26.18 & 42.89 & 93.55 & 28.68 & 37.21 & 94.69 & 26.46 & 40.33 & 94.76 \\
& CoAttack        & 5.96    & 90.55    & 87.28    & 6.01 & 55.64  & 89.82    & 4.32    & 68.49    & 89.42    \\
& \textbf{Ours}   & \textbf{34.55} & \textbf{40.92} & \textbf{94.53} & \textbf{32.65} & \textbf{35.35} & \textbf{95.71} & \textbf{32.54} & \textbf{36.24} & \textbf{95.75} \\
\midrule

\multirow{8}{*}{$\mathrm{DMMI}_{\text{res}}$}
& Attack-SAM      & -- & -- & -- & -- & -- & -- & -- & -- & -- \\
& SRA             & -- & -- & -- & -- & -- & -- & -- & -- & -- \\
& Single-Text     & 43.07 & 23.89 & 94.56 & 27.00 & 29.97 & 94.03 & 35.03 & 16.55 & 95.04 \\
& Multi-Text      & 65.97 & 11.65 & 97.02 & 57.07 & 13.76 & 97.11 & 64.01 & 8.52  & 97.32 \\
& SegPGD-Single   & 34.38 & 25.72 & 92.16 & 20.75 & 31.20 & 92.22 & 25.94 & 18.01 & 92.29 \\
& SegPGD-Multi    & 55.51 & 17.63 & 96.17 & 44.20 & 20.54 & 96.93 & 47.01 & 12.76 & 96.63 \\
& CoAttack        & 10.34    & 28.76    & 89.82    & 9.57    & 25.34    & 88.43    & 10.28    & 13.63    & 95.32    \\
& \textbf{Ours}   & \textbf{71.02} & \textbf{10.24} & \textbf{97.49} & \textbf{63.46} & \textbf{11.56} & \textbf{97.68} & \textbf{70.14} & \textbf{7.26}  & \textbf{97.75} \\
\midrule

\multirow{8}{*}{$\mathrm{DMMI}_{\text{Swin-T}}$}
& Attack-SAM      & -- & -- & -- & -- & -- & -- & -- & -- & -- \\
& SRA             & -- & -- & -- & -- & -- & -- & -- & -- & -- \\
& Single-Text     & 32.83 & 29.92 & 93.41 & 22.18 & 21.83 & 93.58 & 18.97 & 37.99 & 93.14 \\
& Multi-Text      & 63.57 & 13.48 & 96.75 & 56.65 & 9.64 & 97.00  & 50.78 & 14.80 & 96.92 \\
& SegPGD-Single   & 22.78 & 34.52 & 91.23 & 12.71 & 22.21 & 91.84 & 10.02 & 36.98 & 91.27 \\
& SegPGD-Multi    & 47.82 & 21.01 & 95.56 & 39.32 & 14.04 & 95.95 & 60.23 & 12.47 & 96.20 \\
& CoAttack        & 5.34 & 20.66 & 95.56 & 6.82 & 18.39 & 95.64 & 8.02 & 21.95 & 91.77 \\
& \textbf{Ours}   & \textbf{68.14} & \textbf{10.12} & \textbf{97.22} & \textbf{61.22} & \textbf{7.65} & \textbf{97.25} & \textbf{65.56} & \textbf{8.71} & \textbf{97.67} \\
\midrule

\multirow{8}{*}{LAVT}
& Attack-SAM      & -- & -- & -- & -- & -- & -- & -- & -- & -- \\
& SRA             & -- & -- & -- & -- & -- & -- & -- & -- & -- \\
& Single-Text     & 24.31 & 26.61 & 92.76 & 12.49 & 21.13 & 92.13 & 11.70 & 20.29 & 91.63 \\
& Multi-Text      & 52.22 & 12.91 & 96.21 & 43.52 & 10.23 & 96.00 & 39.00 & 11.04 & 95.57 \\
& SegPGD-Single   & 32.99 & 29.70 & 92.68 & 18.79 & 22.05 & 92.44 & 20.60 & 21.84 & 92.43 \\
& SegPGD-Multi    & 55.42 & 12.16 & 96.55 & 43.12 & 10.46 & 95.86 & 42.34 & 10.25 & 96.11 \\
& CoAttack        & 12.35 & 15.73    & 93.49   & 18.32 & 13.65    & 91.78    & 12.23    & 15.06    & 91.44    \\
& \textbf{Ours}   & \textbf{58.16} & \textbf{11.32} & \textbf{96.59} & \textbf{46.75} & \textbf{9.03} & \textbf{96.40} & \textbf{45.16} & \textbf{9.61} & \textbf{96.12} \\
\bottomrule
\end{tabular}
}
\caption{
Quantitative comparison of different adversarial attack methods on five RES models (EVF-SAM, EVF-SAM2, $\mathrm{DMMI}_{\text{res}}$, $\mathrm{DMMI}_{\text{Swin-T}}$, and LAVT) across three datasets (RefCOCO, RefCOCOg, RefCOCO+). Evaluation metrics include ASR@30, mIoU-GT, and mIoU-UTM. $\uparrow$ indicates higher is better, and $\downarrow$ indicates lower is better. The best results for each metric are highlighted in bold.
}
\label{tab:mainexp}
\end{table*}

We conduct a comprehensive evaluation of the proposed PEAT framework under white-box attack scenarios across five RES models (EVF-SAM, EVF-SAM2, DMMI$_{res}$, DMMI$_{Swin\text{-}T}$, and LAVT) and three benchmark datasets (RefCOCO, RefCOCOg, RefCOCO+). For evaluation, we select 750 images from each dataset, covering diverse scenarios close to real-world conditions. In total, our experiments involve more than 15,000 referring expressions, providing a large-scale and statistically reliable basis to ensure the credibility of our results.

Tab.~\ref{tab:mainexp} reports the performance of different attack methods in terms of ASR@30, mIoU-GT, and mIoU-UTM. Overall, PEAT achieves consistent and significant improvements over all baselines across all datasets and model architectures. In particular, PEAT substantially boosts ASR@30, the most direct measure of attack success, while simultaneously decreasing mIoU-GT and increasing mIoU-UTM. These results indicate that PEAT is able to severely disrupt the segmentation capability of RES models while maintaining strong transferability to unseen textual inputs.

When examining the baselines, Multi-Text and SegPGD-Multi perform better than Single-Text and SegPGD-Single by leveraging multiple textual inputs during optimization. However, they still show clear signs of overfitting and limited generalization in more challenging cross-text settings. SegPGD, although effective on conventional segmentation tasks, struggles in RES scenarios due to the multimodal alignment required between vision and language. SRA and Attack-SAM, which rely on point prompts, perform poorly under natural language queries and fail to transfer to RES models because of the inherent modality mismatch between point-based perturbations and language-conditioned segmentation. In addition, we include CoAttack~\citep{coattack}, a representative multimodal attack method. Since CoAttack generates adversarial image–text pairs, we adapt it to our setting by using only the adversarial images as inputs during testing. Despite this adaptation, its results remain consistently weak across all datasets and models, highlighting that paired multimodal perturbations do not translate into effective cross-text adversaries in RES models.

Across models, we also observe that attack performance on EVF-SAM2 is generally lower than on EVF-SAM. This is largely due to the memory mechanism in SAM2~\citep{sam2}, which reinforces predictions using historical visual and prompt embeddings, thereby improving robustness against perturbations. Meanwhile, on DMMI and LAVT, which representative RES models with different architectural backbones, PEAT still demonstrates clear superiority over all baselines, confirming its adaptability and effectiveness across diverse RES designs. Although in some cases baseline methods achieve slightly higher mIoU-UTM (e.g., SegPGD-Multi on certain subsets), the numerical differences are negligible given that background pixels dominate most images and ASR already approaches high levels.

\begin{figure}[!t]
\centering
\includegraphics[width=\linewidth]{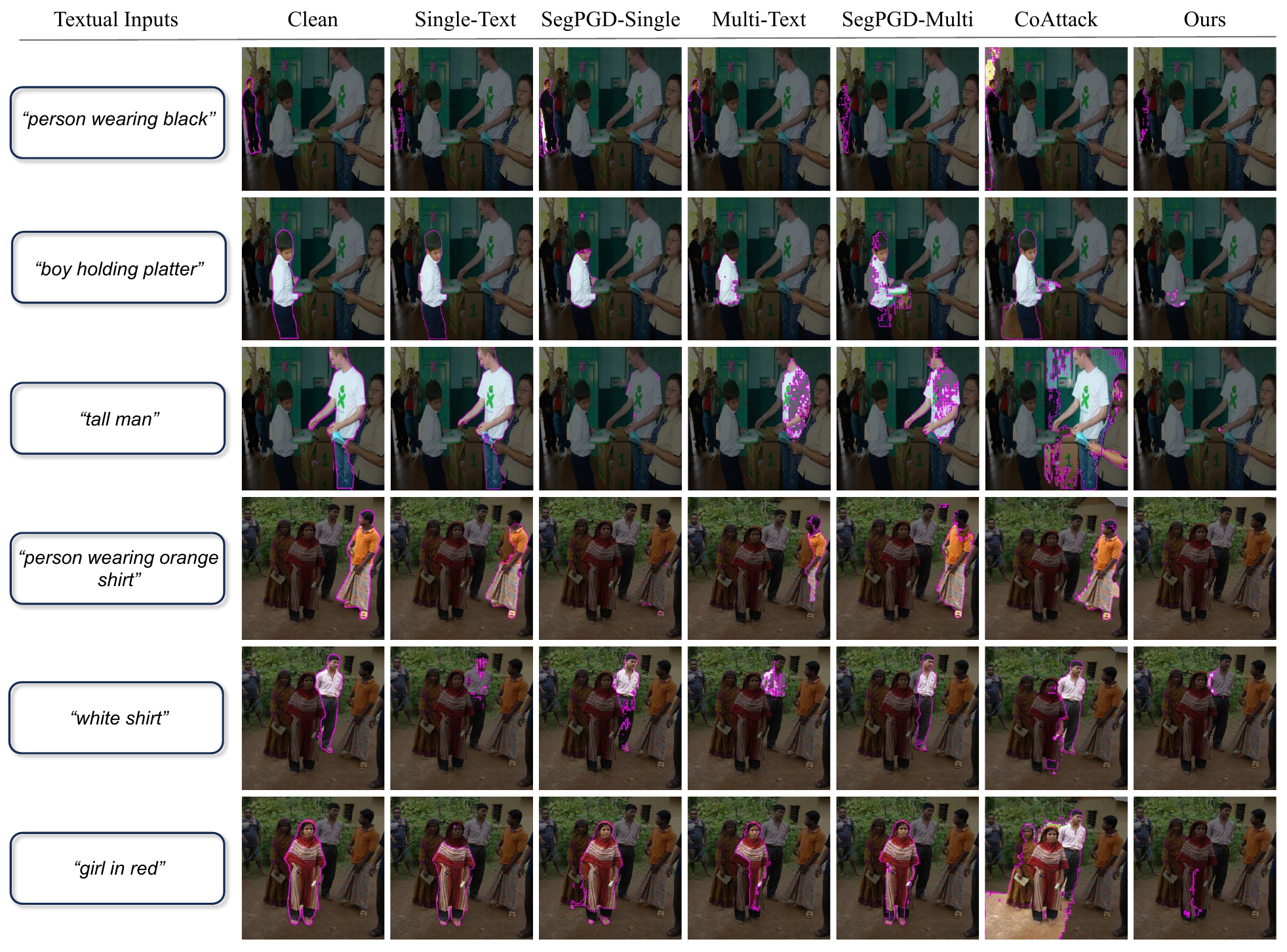}
\caption{Qualitative comparison of adversarial attacks under multiple unseen referring expressions. Each row corresponds to a different textual input, and each column shows the segmentation results under clean visual inputs and different attack methods. Our method consistently suppresses segmentation across diverse unseen expressions, demonstrating superior cross-text generalization compared with baselines.}
\label{fig:vis_comparison}
\end{figure}

Taken together, these results validate the superiority of PEAT in generating adversarial examples with strong attack success and robust cross-text generalization across diverse RES models architectures.

\subsection{Visualization}
We further evaluate the cross-text generalization capability of adversarial images through qualitative comparisons, as shown in Fig.~\ref{fig:vis_comparison}. The figure presents segmentation results under a variety of unseen textual inputs. As observed, single-text methods such as Single-Text and SegPGD-Single exhibit very limited transferability: they are often only effective against the specific textual input used during optimization and fail to generalize to other expressions. Multi-text methods like Multi-Text and SegPGD-Multi show stronger performance by leveraging multiple inputs, but still suffer from overfitting to the optimization texts, resulting in inconsistent effectiveness when tested on unseen descriptions. CoAttack~\citep{coattack}, despite being a representative multimodal attack method that perturbs both image and text, also struggles in the RES setting; since it relies on paired image–text perturbations, its adversarial images alone fail to consistently disrupt segmentation under diverse referring expressions. In contrast, our method consistently produces adversarial images that suppress mask predictions across all unseen textual inputs. This demonstrates that PEAT effectively prevents text-specific overfitting and achieves superior cross-text transferability compared with existing baselines.

\begin{table*}[t]
\centering
\small
\setlength{\tabcolsep}{3pt}
\resizebox{\textwidth}{!}{  
\begin{tabular}{l|ccc|ccc|ccc|ccc}
\toprule
\multirow{2}{*}{Method} & \multicolumn{3}{c|}{$\epsilon_v=4/255$} & \multicolumn{3}{c|}{$\epsilon_v=8/255$} & \multicolumn{3}{c|}{$\epsilon_v=12/255$} & \multicolumn{3}{c}{$\epsilon_v=16/255$} \\
 & ASR@30$\uparrow$ & mIoU-GT$\downarrow$ & mIoU-UTM$\uparrow$ & ASR@30$\uparrow$ & mIoU-GT$\downarrow$ & mIoU-UTM$\uparrow$ & ASR@30$\uparrow$ & mIoU-GT$\downarrow$ & mIoU-UTM$\uparrow$ & ASR@30$\uparrow$ & mIoU-GT$\downarrow$ & mIoU-UTM$\uparrow$  \\
\midrule
Single-Text     & 5.95  & 54.74 & 92.23 & 15.18 & 43.84 & 93.08 & 23.36 & 35.40 & 93.81 & 26.01 & 30.95 & 94.09 \\
Multi-Text      & 9.13  & 46.49 & 93.60 & 27.60 & 29.87 & 95.10 & 44.92 & 18.65 & 96.45 & 57.06 & 13.86 & 97.26 \\
SegPGD-Single   & 3.93  & 55.66 & 91.99 & 11.25 & 42.59 & 92.60 & 14.86 & 35.41 & 92.44 & 22.08 & 31.51 & 92.38 \\
SegPGD-Multi    & 8.59  & 45.81 & 94.28 & 29.19 & 30.33 & 96.03 & 38.75 & 24.22  & 96.57 & 45.22 & 20.59  & 96.97 \\
CoAttack    & 2.38  & 53.69 & 91.68 & 2.43 & 38.33 & 92.11 & 5.73 & 30.26  & 93.57 & 8.94 & 18.36 & 93.73 \\
Ours            & \textbf{10.83} & \textbf{45.04} & \textbf{95.51} & \textbf{30.57} & \textbf{26.99} & \textbf{95.73} & \textbf{48.41} & \textbf{18.58}  & \textbf{96.84} & \textbf{63.46} & \textbf{11.56}  & \textbf{97.68} \\
\bottomrule
\end{tabular}
}  
\caption{Performance under different visual perturbation budgets. For each $\epsilon_v$, we report ASR@30 (\%↑), mIoU-GT (\%↓), and mIoU-UTM (\%↑). Our method consistently outperforms all baselines across various perturbation strengths.}
\label{tab:perturbation_vs_asr}
\vspace{-0.1in}
\end{table*}

\subsection{Performance under Different Visual Perturbation Budgets and ASR Threshold}

\begin{wrapfigure}{r}{0.5\textwidth}
    \centering
    \vspace{-10pt}  
    \includegraphics[width=0.48\textwidth]{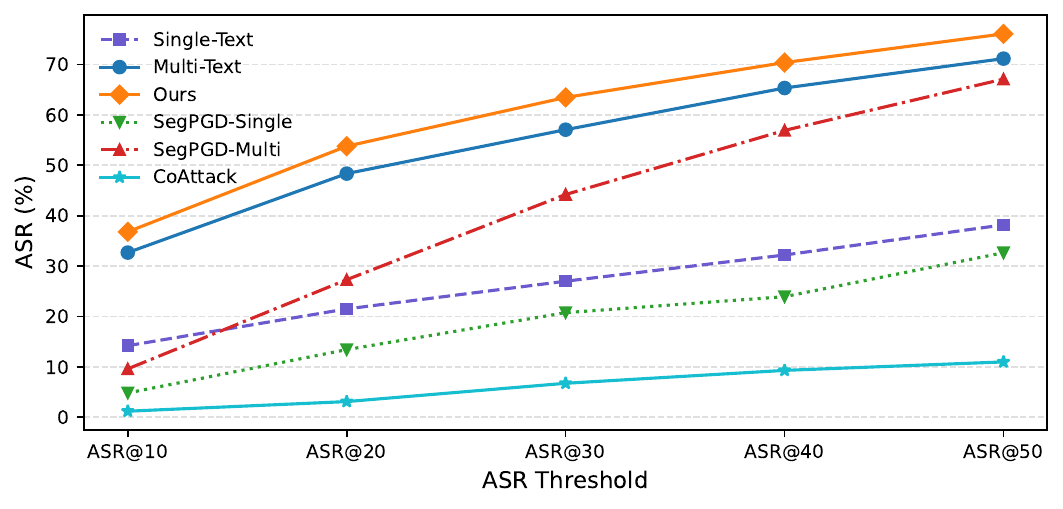}
    \caption{Attack success rates (ASR) of different methods under varying ASR thresholds (ASR@10--ASR@50). Our method consistently achieves the highest ASR across all thresholds.}
    \label{fig:threshold_vs_asr}
    \vspace{-1pt}  
\end{wrapfigure}

To further assess the stability of \textbf{PEAT}, we compare methods under varying attack settings. Table~\ref{tab:perturbation_vs_asr} reports ASR@30, mIoU-GT and mIoU-UTM for four visual perturbation budgets $\epsilon_v\in\{4/255,8/255,12/255,16/255\}$. As expected, performance generally improves as $\epsilon_v$ increases. Importantly, PEAT consistently achieves the best results across all budgets, and its advantage is particularly evident under smaller perturbations (e.g., $\epsilon_v=4/255$ and $\epsilon_v=8/255$), showing that it can produce highly effective adversarial examples even with limited perturbation strength.

Figure~\ref{fig:threshold_vs_asr} further evaluates attack success under different ASR thresholds (ASR@10, ASR@20, ASR@30, ASR@40, ASR@50). Across all tolerance levels, PEAT maintains the highest ASR, with the gap widening under stricter thresholds. While other baselines also improve as the threshold loosens, they remain consistently weaker. 

Overall, these results confirm that PEAT not only delivers strong attack performance under generous budgets and relaxed thresholds, but also retains robustness in more constrained settings, thereby validating its stability and cross-text generalization capability.

\subsection{Ablation Study}

\subsubsection{Effect of proxy embedding perturbation update location}

\begin{wraptable}{13}{0.46\textwidth}
    \centering
    \vspace{-10pt}
    \small
    \setlength{\tabcolsep}{4pt}
    
    \resizebox{0.98\linewidth}{!}{%
    \begin{tabular}{lccc|ccc}
        \toprule
        Model & PEU & MIU & TEU & ASR@30$\uparrow$ & mIoU-GT$\downarrow$ & mIoU-UTM$\uparrow$ \\
        \midrule
        \multirow{3}{*}{EVF-SAM} 
        & \checkmark & & & \textbf{84.06} & \textbf{9.85} & \textbf{98.65} \\
        & & \checkmark & & 83.75 & 10.05 & 98.61 \\
        & & & \checkmark & 80.58 & 11.77 & 98.37 \\
        \midrule
        \multirow{3}{*}{EVF-SAM2} 
        & \checkmark & & & 31.42 & 43.68 & 93.61 \\
        & & \checkmark & & \textbf{34.55} & \textbf{40.92} & \textbf{93.92} \\
        & & & \checkmark & 27.11 & 47.71 & 93.97 \\
        \bottomrule
    \end{tabular}
    }
    \caption{Ablation study on proxy embedding perturbation update locations. PEU is Prompt Embedding Update; MIU is Multimodal Input Update; TEU is Text Embedding Update. PEU is best for EVF-SAM; MIU for EVF-SAM2.}
    \label{tab:ablation_update_location}
    \vspace{-10pt}
\end{wraptable}

We investigate the optimal proxy embedding perturbation update locations for the \textbf{PEAT} by designing three different update strategies on the SAM-based RES models:
1) updating only the prompt embedding in the SAM Prompt Encoder (Prompt Embedding Update, PEU);
2) simultaneously updating both the text embedding and visual inputs in the multimodal encoder (Multimodal Input Update, MIU);
3) updating only the text embedding without modifying the visual inputs (Text Embedding Update, TEU).

It is worth noting that since SAM-based RES models such as EVF-SAM and EVF-SAM2 have distinct architectures that include both a SAM Prompt Encoder and a multimodal encoder, unlike general RES models, we conduct this ablation study exclusively on SAM-based models to comprehensively analyze the impact of different update locations on attack effectiveness.

As shown in Tab.~\ref{tab:ablation_update_location}, we evaluate the three update strategies on EVF-SAM and EVF-SAM2. Experimental results demonstrate that for EVF-SAM, updating perturbations on the prompt embedding (PEU) achieves the best performance, reaching an ASR@30 of 84.06\%, with the lowest mIoU-GT of 9.85, misleading that the model fails to segment the correct target regions. In contrast, both MIU and TEU strategies lead to lower performance.

Interestingly, for EVF-SAM2, the optimal results are achieved under the MIU setting, where textual modality perturbations are applied simultaneously to both textual and visual inputs. Under this strategy, the ASR@30 reaches 34.55\%, significantly higher than that achieved by TEU alone. This observation may be attributed to differences in the feature fusion mechanisms between EVF-SAM and EVF-SAM2, where updating only the text embedding is insufficient to effectively disrupt the final segmentation outputs.

Overall, these findings suggest that selecting appropriate perturbation update locations according to model architectures is crucial for maximizing cross-text transferability in adversarial attacks.

\subsubsection{Effect of perturbation update interval}
\label{interval}
We first study how the update interval $N$ for proxy embedding perturbation affects adversarial performance. In our optimization, the proxy embedding is updated every $N$ steps rather than at each iteration. As shown in Tab.~\ref{tab:update_interval}, ASR@30 improves as $N$ increases and consistently peaks at $N=10$ across all models: 83.99\% on EVF-SAM, 34.51\% on EVF-SAM2, 70.96\% on DMMI$_{res}$, 68.22\% on DMMI$_{Swin\text{-}T}$, and 58.16\% on LAVT. This setting also yields the lowest mIoU-GT and highest mIoU-UTM on most models (e.g., mIoU-GT drops to 10.28 on DMMI$_{res}$ and 11.30 on LAVT, while mIoU-UTM reaches 97.36 and 96.61, respectively). Both overly frequent updates ($N=1$) and overly sparse updates ($N\geq15$) lead to reduced effectiveness, suggesting that a moderate interval balances alignment and generalization. Therefore, we adopt $N=10$ as the default configuration.

\subsubsection{Effect of proxy embedding perturbation step size}
\label{step}
We also evaluate the influence of the step size $\alpha_t$ for updating the proxy embedding, with perturbation budget fixed at 0.2 and 10 iterations. Results in Tab.~\ref{tab:step_size} show that $\alpha_t=0.01$ consistently achieves the best ASR@30 across all models: 84.06\% on EVF-SAM, 34.55\% on EVF-SAM2, 71.02\% on DMMI$_{res}$, 68.13\% on DMMI$_{Swin\text{-}T}$, and 58.69\% on LAVT. This setting also corresponds to the lowest mIoU-GT (9.85, 40.92, 10.24, 10.09, and 11.34, respectively) and competitive or highest mIoU-UTM values. Larger step sizes degrade performance, likely due to unstable optimization trajectories, whereas smaller and smoother updates promote stable convergence under limited iterations.

\begin{table*}[t]
\centering
\small
\begin{minipage}[t]{0.49\linewidth}
\centering
\setlength{\tabcolsep}{3pt}
\resizebox{\linewidth}{!}{
\begin{tabular}{llccccc}
\toprule
\textbf{Model} & \textbf{Metric} & \textbf{1} & \textbf{5} & \textbf{10} & \textbf{15} & \textbf{20} \\
\midrule
\multirow{3}{*}{EVF-SAM}
 & ASR@30  & 78.57 & 83.87 & \textbf{83.99} & 82.45 & 82.04 \\
 & mIoU-GT $\downarrow$ & 12.43 & 9.16 & \textbf{9.54} & 10.36 & 10.35 \\
 & mIoU-UTM $\uparrow$ & 98.25 & 98.55 & \textbf{98.61} & 98.48 & 98.44 \\
\midrule
\multirow{3}{*}{EVF-SAM2}
 & ASR@30  & 32.38 & 33.99 & \textbf{34.51} & 33.23 & 32.84 \\
 & mIoU-GT $\downarrow$ & 41.52 & 41.00 & \textbf{40.95} & 41.84 & 41.92 \\
 & mIoU-UTM $\uparrow$ & 93.98 & 94.48 & \textbf{94.50} & 93.59 & 93.32 \\
\midrule
\multirow{3}{*}{$\mathrm{DMMI}_{\text{res}}$}
 & ASR@30  & 68.94 & 70.44 & \textbf{70.96} & 70.56 & 70.10 \\
 & mIoU-GT $\downarrow$ & 12.35 & 12.13 & \textbf{10.28} & 12.03 & 10.98 \\
 & mIoU-UTM $\uparrow$ & 95.21 & 95.89 & \textbf{97.36} & 97.29 & 96.73 \\
\midrule
\multirow{3}{*}{$\mathrm{DMMI}_{\text{Swin-T}}$}
 & ASR@30  & 63.88 & 62.49 & \textbf{68.22} & 68.03 & 67.37 \\
 & mIoU-GT $\downarrow$ & 11.38 & 10.65 & \textbf{10.08} & 10.23 & 11.49 \\
 & mIoU-UTM $\uparrow$ & 97.01 & 96.98 & 97.21 & \textbf{97.25} & 96.48 \\
\midrule
\multirow{3}{*}{LAVT}
 & ASR@30  & 55.32 & 57.99 & \textbf{58.16} & 58.08 & 57.65 \\
 & mIoU-GT $\downarrow$ & 12.63 & 12.14 & \textbf{11.30} & 11.94 & 11.98 \\
 & mIoU-UTM $\uparrow$ & 95.86 & 95.79 & \textbf{96.61} & 96.14 & 95.34 \\
\bottomrule
\end{tabular}
}
\caption{Ablation study on proxy embedding perturbation update interval $N$. Best results occur at $N=10$.}
\label{tab:update_interval}
\end{minipage}
\hfill
\begin{minipage}[t]{0.49\linewidth}
\centering
\setlength{\tabcolsep}{2pt}
\begin{tabular}{llrrrrr}
\toprule
Model & Metric & 0.01 & 0.02 & 0.03 & 0.04 & 0.05 \\
\midrule
\multirow{3}{*}{EVF-SAM} 
  & ASR@30         & \textbf{84.06} & 83.76 & 82.02 & 80.64 & 79.17 \\
  & mIoU-GT$\downarrow$ & \textbf{9.85} & 9.55 & 10.83 & 11.25 & 12.87 \\
  & mIoU-UTM$\uparrow$   & 98.61 & \textbf{98.65} & 97.22 & 97.99 & 98.22 \\
\midrule
\multirow{3}{*}{EVF-SAM2} 
  & ASR@30         & \textbf{34.55} & 32.94 & 32.13 & 31.99 & 31.42 \\
  & mIoU-GT$\downarrow$ & \textbf{40.92} & 42.15 & 41.23 & 43.11 & 43.68 \\
  & mIoU-UTM$\uparrow$   & \textbf{94.53} & 93.78 & 94.16 & 94.59 & 93.63 \\
\midrule
\multirow{3}{*}{$\mathrm{DMMI}_{\text{res}}$}
  & ASR@30         & \textbf{71.02} & 70.57 & 69.58 & 69.12 & 67.64 \\
  & mIoU-GT$\downarrow$ & \textbf{10.24} & 11.46 & 11.99 & 11.23 & 15.35 \\
  & mIoU-UTM$\uparrow$   & \textbf{97.49} & 97.12 & 96.26 & 97.03 & 95.33 \\
\midrule
\multirow{3}{*}{$\mathrm{DMMI}_{\text{Swin-T}}$}
  & ASR@30         & \textbf{68.13} & 68.02 & 67.86 & 67.88 & 67.23 \\
  & mIoU-GT$\downarrow$ & \textbf{10.09} & 10.13 & 10.88 & 11.92 & 11.85 \\
  & mIoU-UTM$\uparrow$   & \textbf{97.22} & 97.05 & 97.06 & 96.87 & 96.75 \\
\midrule
\multirow{3}{*}{LAVT} 
  & ASR@30         & \textbf{58.69} & 57.61 & 57.38 & 56.98 & 56.83 \\
  & mIoU-GT$\downarrow$ & \textbf{11.34} & 11.59 & 11.64 & 12.01 & 12.11 \\
  & mIoU-UTM$\uparrow$   & \textbf{96.60} & 96.44 & 95.99 & 96.04 & 95.86 \\
\bottomrule
\end{tabular}
\caption{Ablation study on step size $\alpha_t$ for proxy embedding perturbation. Best results occur at $\alpha_t=0.01$.}
\label{tab:step_size}
\end{minipage}
\end{table*}

These ablation results collectively validate the effectiveness of our proposed design choices in enhancing attack effectiveness and cross-text transferability. They further highlight the importance of precise update locations, appropriately spaced optimization intervals, and appropriate step sizes in \textbf{PEAT} for RES models.

\section{Conclusion}
\label{sec:conclu}
In this paper, we presented a systematic study on the adversarial robustness of Referring Expression Segmentation (RES) models, a critical yet underexplored problem in multimodal vision–language understanding. To address the unique challenges arising from multimodal interactions, we introduced \textbf{Embedding-Guided Bidirectional Attack (PEAT)}, a novel embedding-guided bidirectional attack framework. PEAT integrates two complementary components: \textbf{Visual-Aligned Optimization} (VAO), which drives adversarial images toward a predefined target mask, and \textbf{Embedding-Guided Adversarial Optimization} (EGAO), which adversarially updates a proxy embedding to misalign the textual branch and guide visual perturbations toward more generalizable adversarial directions. This design enables the generation of adversarial images that remain effective across diverse and unseen textual inputs. 

Extensive experiments across multiple benchmark datasets and diverse RES architectures demonstrate that PEAT consistently surpasses existing baselines in terms of attack success, cross-text transferability, and robustness under varying perturbation strengths. Qualitative visualizations further illustrate its ability to suppress segmentation across different textual inputs, and ablation studies validate the necessity of each design choice. 

Overall, our work not only reveals the latent vulnerabilities of RES models to cross-text adversarial attacks but also provides a foundation for future research on robust and trustworthy RES systems. In future work, we will extend PEAT to black-box settings, investigate cross-model transferability in RES models, and design interpretable and controllable defense mechanisms tailored for this task.

\section{Acknowledgment}
This work is supported by the Yunnan Province Expert Workstations (Grant No: 202305AF150078), the National Natural Science Foundation of China (Grant No: 62162067), the Yunnan Fundamental Research Project (Grant Nos: 202401AT070474, 202501AU070059), the Yunnan Province Special Project (Grant No: 202403AP140021), the Yunnan Provincial Department of Education Science Research Project (Grant No: 2025J0006), and the Fourth Professional Degree Graduate Practice Innovation Project of Yunnan University (Grant Nos: ZC-24249746).


\bibliographystyle{elsarticle-num} 
\bibliography{refs}






\end{document}